\documentclass{article}

\PassOptionsToPackage{numbers}{natbib}
\usepackage[preprint]{neurips_2021}

\usepackage[utf8]{inputenc} 
\usepackage[T1]{fontenc}    
\usepackage[pagebackref=true,breaklinks=true,colorlinks,bookmarks=false]{hyperref}       
\usepackage{url}            
\usepackage{booktabs}       
\usepackage{amsfonts}       
\usepackage{nicefrac}       
\usepackage{microtype}      
\usepackage[dvipsnames, table, x11names]{xcolor}         
\usepackage{amsmath}
\usepackage{graphicx}
\usepackage{makecell}
\usepackage{multirow}
\usepackage{caption}
\usepackage{comment}
\usepackage{enumitem}
\usepackage{float}

\newcommand{\todo}[1]{\textcolor{Red}{#1}}%
\newcommand{\gain}[1]{\textcolor{Green}{(+\textbf{#1})}}%
\newcommand{\ignore}[1]{\textcolor{Gray}{(#1)}}%

\title{Visual Parser: Representing Part-whole Hierarchies with Transformers}

\author{%
  Shuyang Sun$^{*\dagger}$, Xiaoyu Yue$^*$, Song Bai$^\ddagger$, Philip Torr$^\dagger$
    \\
  $^\dagger$University of Oxford, $^\ddagger$ByteDance AI Lab\\
  \texttt{\{kevinsun, phst\}@robots.ox.ac.uk, \{yuexiaoyu002, songbai.site\}@gmail.com} \\
}

\begin{document}

\maketitle

\begin{abstract}
  Human vision is able to capture the part-whole hierarchical information from the entire scene. This paper presents the  Visual Parser (ViP) that explicitly constructs such a hierarchy with transformers. ViP divides visual representations into two levels, the \emph{part} level and the \emph{whole} level. Information of each part represents a combination of several independent vectors within the whole. To model the representations of the two levels, we first encode the information from the whole into part vectors through an attention mechanism, then decode the global information within the part vectors back into the whole representation. By iteratively parsing the two levels with the proposed encoder-decoder interaction, the model can gradually refine the features on both levels. Experimental results demonstrate that ViP can achieve very competitive performance on three major tasks \textit{e.g.} classification, detection and instance segmentation. In particular, it can surpass the previous state-of-the-art CNN backbones by a large margin on object detection. The tiny model of the ViP family with $7.2\times$ fewer parameters and $10.9\times$ fewer FLOPS can perform comparably with the largest model ResNeXt-101-64$\times$4d of ResNe(X)t family. 
  Visualization results also demonstrate that the learnt parts are highly informative of the predicting class, making ViP more explainable than previous fundamental architectures.
  Code is available at \url{https://github.com/kevin-ssy/ViP}.
  
\end{abstract}

\section{Introduction}
Strong evidence has been found in psychology that human vision is able to parse a complex scene into part-whole hierarchies with many different levels from the low-level pixels to the high-level properties (\textit{e.g.} parts, objects, scenes) \citep{partwhole_evidence1, partwhole_evidence2}.
 Constructing such a part-whole hierarchy enables neural networks to capture compositional representations directly from images, which can promisingly help to detect properties of many different levels with only one network. 
 
 To the best of our knowledge, most current visual feature extractors do not model such hierarchy explicitly.
 Due to the lack of such part-whole hierarchy in representation modeling, existing feature extractors cannot find the compositional features directly from the network. 
 Ideal modeling of the visual representation should be able to model the part-whole hierarchy as humans do so that we can leverage representations of all levels directly from one backbone model.
 
 Building up a framework that includes different levels of representations in the part-whole hierarchy is difficult for conventional neural networks as it requires neurons to dynamically respond to the input, while neural networks with fixed weights cannot dynamically allocate a group of neurons to represent a node in a parse tree \citep{glom}. With the rise of the Transformer \citep{transformer}, such a problem can be possibly resolved due to its dynamic nature.
 \begin{figure}
    \centering
    \includegraphics[scale=0.66]{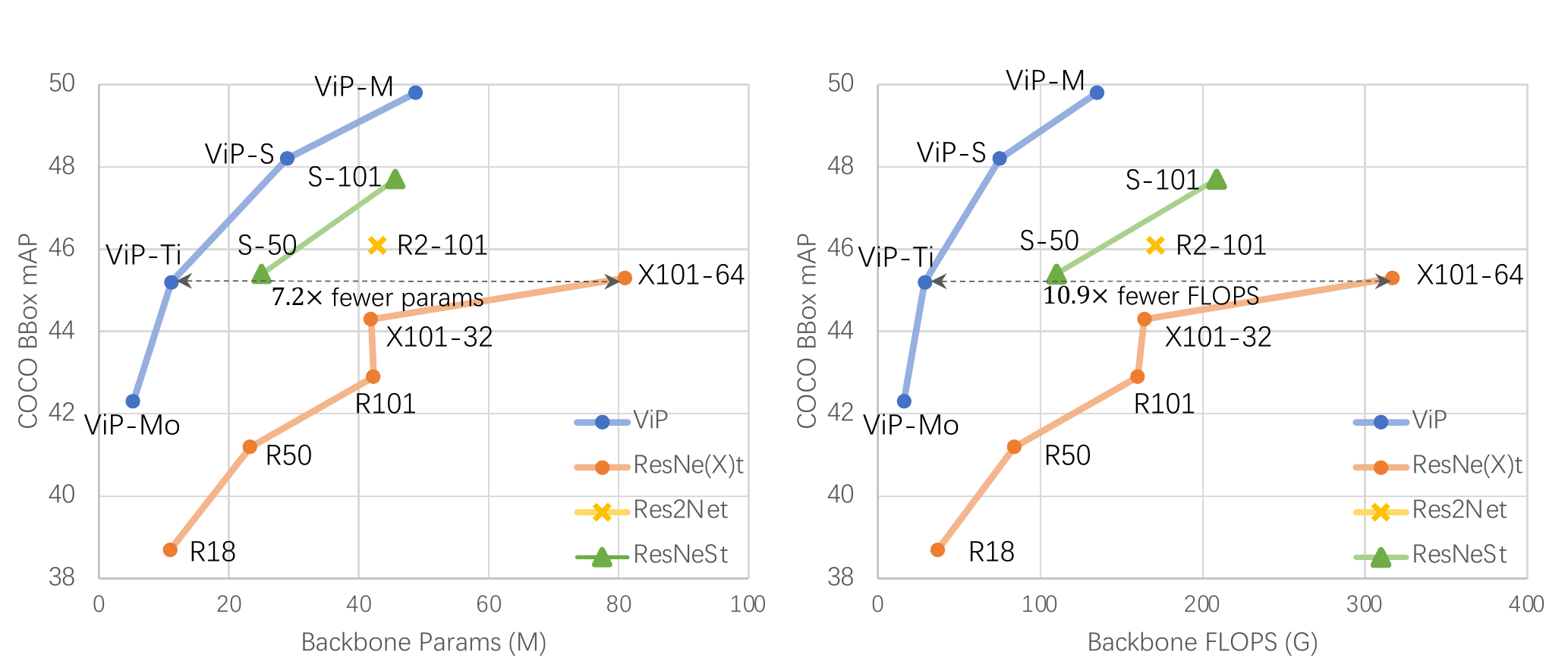}
    \caption{\textbf{ViP can outperform all state-of-the-art CNN backbones by a large margin on object detection.}  All listed results are trained under the regular $1\times$ regime in Cascade Mask-RCNN \citep{cascade-rcnn}.
    }
    \label{fig:det_performance}
\end{figure}
 
 In this paper, we show how to construct a simplest part-whole hierarchy for visual representation. The hierarchy of the proposed network has two levels. One represents the \textit{part}, which only contains the most essential information describing the visual input, and the other is for the \textit{whole}, which describes the visual input in a regular spatial coordinate frame system. Normally a part vector can be dynamically associated with several vectors of the whole, forming a one-to-many mapping between the two levels.
 To obtain information for the part, we first apply an encoder between the two levels to fill each part with the features of the whole. Then the encoded part feature will be mapped back to the whole by a transformer-based decoder. 
 Such cross-level interaction is iteratively applied throughout the network, constituting a bi-directional pathway between the two levels. 
 
 \textbf{Our main contributions} are as follows: 
 (1) We propose Visual Parser (ViP) that can explicitly construct a part-whole hierarchy within the network for basic visual representations. This is the first step towards learning multi-level part-whole representations. 
 (2) With the help of such a part-whole hierarchy, the network can be more explainable compared with previous networks.
 (3) ViP can be directly applied as a backbone for versatile uses. Experimental results also demonstrate that ViP can achieve very competitive results compared to the current state-of-the-art backbones. As shown in Figure \ref{fig:det_performance}, it outperforms all state-of-the-art CNN counterparts by a large margin on object detection.

 \section{Method}
 \label{sec:method}
 \subsection{Overview}
\begin{figure}
    \centering
    \includegraphics[scale=0.366]{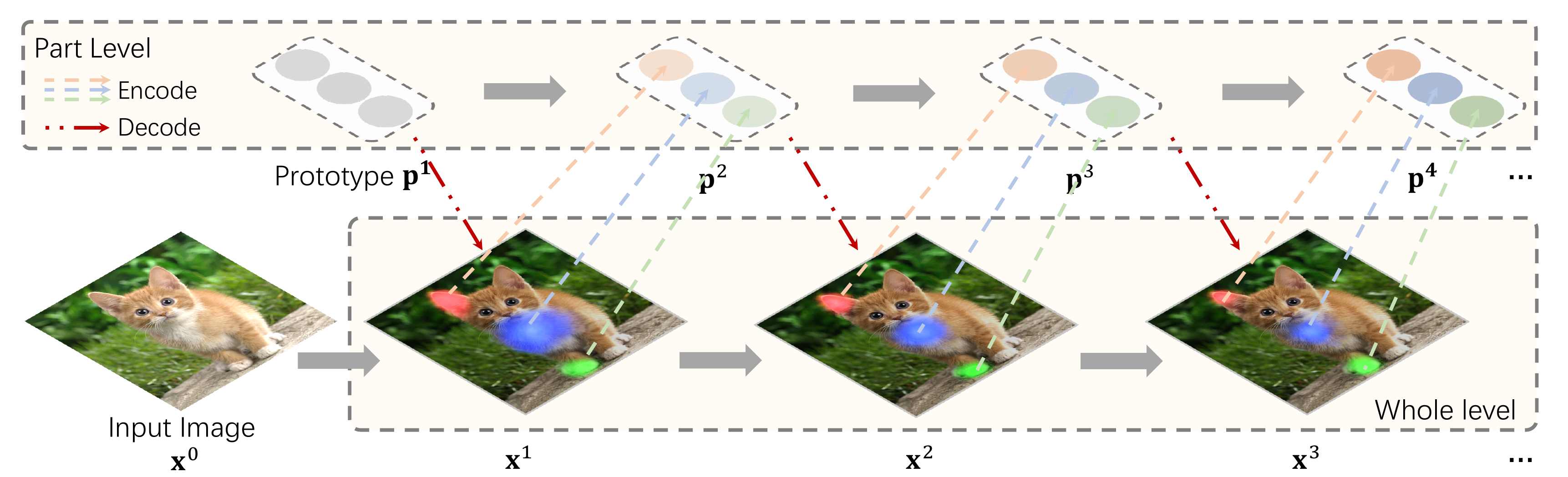}
    \caption{The overall pipeline of the Visual Parser (ViP). Given an input image and the prototype parameters $\mathbf{p}^1$ as input, we iteratively refine the features of the two levels with the proposed encoder-decoder process.}
    \label{fig:overview}
\end{figure}

 The overall pipeline is shown in Figure \ref{fig:overview}. There are two inputs of ViP, including an input image and a set of learnable parameters. These parameters represent the prototype of the parts, and will be used as initial clues indicating the region that each part should associate with.
 The entire network consists of several basic blocks (iterations). For block $i \in \{2, 3, ..., B\}$, there are two kinds of representations describing the two hierarchical levels. One is the part representation $\mathbf{p}^i \in \mathbb{R}^{N\times C}$ and the other is the whole feature maps $\mathbf{x}^i \in \mathbb{R}^{L\times C}$. Here $N$ is a pre-defined constant number indicating the number of parts within the input image, and $L$ is the number of pixels of the feature map, which is identical to \textit{width$\times$height}. $B, C$ are the numbers of blocks and channels respectively.
 The representation of parts for each block is dynamically encoded from the corresponding whole feature maps through an attention-based approach. Given the representation of the part $\mathbf{p}^{i-1}$ and the whole $\mathbf{x}^{i-1}$ from the previous block $i-1$, an attention-based encoder is applied to fill the information of the whole into the part $\mathbf{p}^{i}$ of the current block. Since the attention mechanism assigns each pixel on the feature map with a weight indicating the affinity between the pixel and the corresponding part, only the spatial information in $\mathbf{x}^{i-1}$ that is semantically close to the input part $\mathbf{p}^{i-1}$ can be updated into $\mathbf{p}^{i}$.
 
 Information within the encoded parts will be then transferred back into the feature maps, so each pixel on the feature map can interact with the information in a wider range. Since the information within the parts is highly condensed, the computational cost  between the pixels and the parts is much lower than the original pixel-wise global attention \citep{nonlocal, botnet}. This encoder-decoder process constitutes the basic building block of ViP. By stacking the building block iteratively, the network can learn to construct a two-level part-whole hierarchy explicitly.
 
 \subsection{Part Encoder}
 The part encoder is responsible for extracting the part information based on the previous part-whole input. The encoder is implemented with an attention mechanism.
 Given the part representation $\mathbf{p}^{i-1} \in \mathbb{R}^{N\times C}$ of the last block $i-1$, we first normalize it with Layer Normalization \citep{layernorm}, then use it as the query of the attention block. The whole feature map from the last block $\mathbf{x}^{i-1} \in \mathbb{R}^{L\times C}$ serves as the key and value after the normalization. The information of the whole will be condensed into the part representations via attention, which can be formulated as:
 \begin{equation}
 \begin{split}
     &\mathbf{\hat{p}}^{i-1} = \mathbf{p}^{i-1} + \text{Attention}(\mathbf{p}^{i-1}+\mathbf{d}_e, \mathbf{x}^{i-1} + \mathbf{d}_w, \mathbf{x}^{i-1}),\\
 \end{split}
 \label{eq:attn_def}
 \end{equation}
 where $\mathbf{\hat{p}}^{i-1} \in \mathbb{R}^{N\times C}$ is the output of the attention block, and $\text{Attention}(\textit{query}, \textit{key}, \textit{value})$ denotes the attention mechanism. $\mathbf{d}_w \in \mathbb{R}^{L\times C}, \mathbf{d}_e \in \mathbb{R}^{N\times C}$ are positional encodings for the whole and the part respectively. The positional encoding $\mathbf{d}_w$ follows the sinusoidal design proposed in \citep{detr}, and $\mathbf{d}_e$ is a set of learnable weights.
 We follow the common practice of the classic attention calculation, which first outputs an affinity matrix $\mathbf{M} \in \mathbb{R}^{N\times L}$ between the query and key and then use it to select information lying in the value. 
 \begin{equation}
     \mathbf{M} = \frac{1}{\sqrt{C}} q(\text{LN}(\mathbf{p}^{i-1}+\mathbf{d}_e)) \cdot k(\text{LN}(\mathbf{x}^{i-1} + \mathbf{d}_w))^T,
     \label{eq:softmax}
 \end{equation}
 where the functions $q(\cdot), k(\cdot)$ denote the linear mappings for the inputs of query and key, $\text{LN}$ is the Layer Normalization. Here we omit the learnable weights within LN for simplicity. The product of the query and the key is normalized by a temperature factor $\frac{1}{\sqrt{C}}$ to avoid it being one-hot after the lateral softmax calculation. 
 The softmax operation guarantees the sum of the affinity matrix on the whole dimension to be one, which can be formulated as:
 \begin{equation}
     \hat{M}_{a, b} = \frac{e^{M_{a, b}}}{\sum_l{e^{{M}_{a, l}}}}.
 \end{equation}
 The normalized affinity matrix can be easy to explain, as it assigns an independent weight to each spatial location indicating where each part is lying on the spatial feature maps. To aggregate these weighted spatial locations into part vectors, we follow the classic attention that weighted averaging the values together with the affinity matrix:
 \begin{equation}
     \text{Attention}(\mathbf{p}^{i-1}+\mathbf{d}_e, \mathbf{x}^{i-1} + \mathbf{d}_w, \mathbf{x}^{i-1}) = \hat{\mathbf{M}} \cdot v(\text{LN}(\mathbf{x}^{i-1})),
 \end{equation}
 where $v(\cdot)$ is the linear mapping for values, and $\hat{\mathbf{M}}$ is the affinity matrix after softmax.
 
 \textbf{Reasoning across different parts.} After each part is filled with the information from the feature maps, we apply a part-wise reasoning module to enable information communication between parts. In order to save computational cost, we just apply a single linear projection with learnable weights $\mathbf{W}_{p} \in \mathbb{R}^{N\times N}$. An identity mapping and the normalization are also applied here as a residual block. The process of the part-wise reasoning can be formulated as:
 \begin{equation}
     \mathbf{\hat{p}}_r^{i-1} = \mathbf{\hat{p}}^{i-1} + \mathbf{W}_{p} \cdot \text{LN}(\mathbf{\hat{p}}^{i-1}),
 \end{equation}
 where $\mathbf{\hat{p}}_r^{i-1}$ represents the output for the part-wise reasoning.
 
 \textbf{Activating the part representations.} The part representation learnt above may not be all meaningful since different objects may have different numbers of parts describing themselves. We thereby further apply a Multi-Layer Perceptron (MLP) that has two linear mappings with weight $\mathbf{W}_{f1}, \mathbf{W}_{f2} \in \mathbb{R}^{C\times C}$ and an activation function (GELU \citep{gelu}) $\sigma(\cdot)$ in its module. The activation function will only keep the useful parts to be active, while those identified to be less helpful will be squashed.
 In this way, we obtain the part representation $\mathbf{p}^{i}$ for block $i$ by:
 \begin{equation}
     \begin{split}
        \mathbf{p}^{i} &= \mathbf{\hat{p}}_r^{i-1} + \text{MLP}(\mathbf{\hat{p}}_r^{i-1}),\\
        \text{MLP}(\mathbf{\hat{p}}_r^{i-1}) &= \sigma(\text{LN}(\mathbf{\hat{p}}_r^{i-1}) \cdot \mathbf{W}_{f1}) \cdot \mathbf{W}_{f2}.
     \end{split}
\label{eq:rpn_mlp}
 \end{equation}
The above process demonstrates that the part representation generated by the previous block will be used to initialize the parts of the next iteration. Thus the randomly initialized part representations will be gradually refined with the information from the whole within each block.

\begin{figure}[t]
\begin{minipage}{\textwidth}
    \begin{minipage}{0.58\textwidth}
        \includegraphics[scale=0.333]{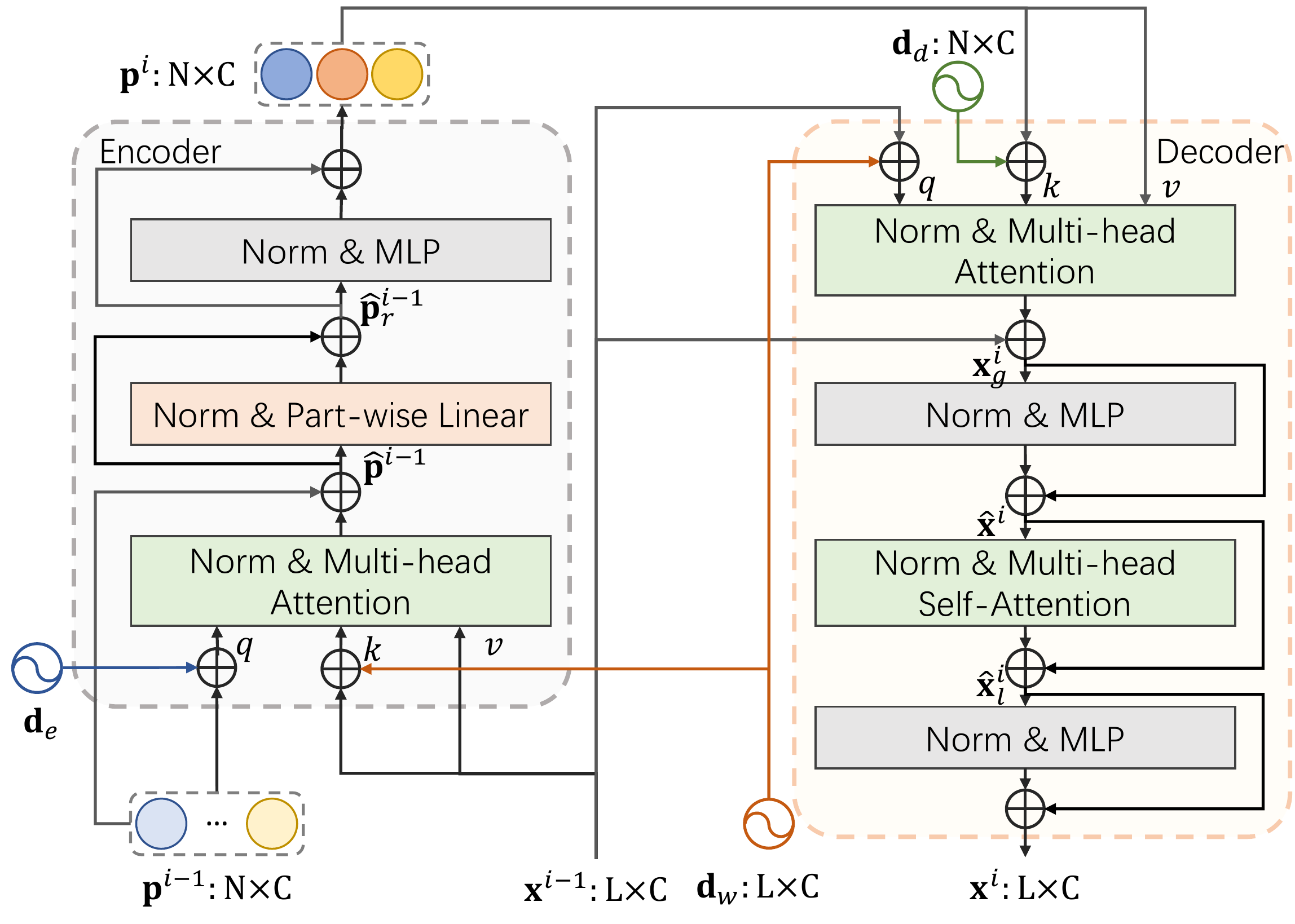}
        \caption{The basic building block of ViP. The symbol $\bigoplus$ denotes element-wise summation. Here we omit the relative positional embedding $\mathbf{r}^i$ for clear demonstration.}
    \label{fig:block}
    \end{minipage}
    \hfill\hfill
    \begin{minipage}{0.38\textwidth}
        \setlength{\tabcolsep}{6.6pt}
        \renewcommand{\arraystretch}{1.5}
        \centering
        \scalebox{0.76}{
        \begin{tabular}{c|c|c}
        \Xhline{1.5pt}
            Stage & L & \makecell[c]{Block Settings}\\
            \hline
             & & \makecell[c]{$7\times7, 64$, Conv, stride 2 \\ $3\times 3$ Max Pool, stride 2} \\
            \hline
            \multirow{2}{*}{Stage$_1$} & \multirow{2}{*}{$\frac{HW}{4^2}$}
            & Patch Embedding, $C_1$ \\ \cline{3-3}
            & & \makecell[c]{
            $\begin{bmatrix} C_1, N_1, G_1  \end{bmatrix} \times B_1$} \\
            \hline
            \multirow{2}{*}{Stage$_2$} & \multirow{2}{*}{$\frac{HW}{8^2}$}
            & Patch Embedding, $C_2$ \\ \cline{3-3}
             & & $[ C_2, N_2, G_2 ] \times B_2$ \\
            \hline
            \multirow{2}{*}{Stage$_3$} & \multirow{2}{*}{$\frac{HW}{16^2}$}
            & Patch Embedding, $C_3$ \\ \cline{3-3} 
            & & \makecell[c]{
            $\begin{bmatrix} C_3, N_3, G_3  \end{bmatrix} \times B_3$} \\
            \hline
            \multirow{2}{*}{Stage$_4$} & \multirow{2}{*}{$\frac{HW}{32^2}$}
            & Patch Embedding, $C_4$ \\ \cline{3-3}
            & & \makecell[c]{
            $\begin{bmatrix} C_4, N_4, G_4  \end{bmatrix} \times B_4$}\\
            \hline
             &  & \makecell[c]{Last Encoder: [$C_4, N_4, G_4$] \\ $1000$, Linear}\\
            \Xhline{1.5pt}
        \end{tabular}
        }
        \captionof{table}{General specification for ViP family. Building blocks of ViP are shown in brackets. The last encoder is only set for ViP-Mo/Ti/S.}
        \label{tab:table_net}
    \end{minipage}
\end{minipage}
\end{figure}
\subsection{Whole Decoder}
As shown in Figure \ref{fig:block}, there are two inputs for the decoder, the part representation $\mathbf{p}^{i}$ and the whole representation $\mathbf{x}^{i-1}$. Interactions within the decoder can be divided into a part-whole global interaction between the parts and the feature maps, and a patch-based local attention between pixels in a local window of the whole feature maps.

\textbf{Part-whole global interaction.} We first apply the part-whole global attention to fill each pixel on the whole representations $\mathbf{x}^{i-1}$ with the global information encapsulated in the parts $\mathbf{p}^{i}$. The part-whole global attention completely follows the classic attention paradigm \citep{transformer}, which takes $\mathbf{x}^{i-1}$ as the query input, and $\mathbf{p}^{i}$ as inputs of the key and value. Therefore each pixel of the whole representation can have a long-range interaction with the encoded parts. Before feeding into the attention, both part and whole representations will be normalized by Layer Normalization. An identity mapping and a MLP are also applied as what does in common practice.
The process of the part-whole interaction in the decoder can be written as:
\begin{equation}
\begin{split}
    \mathbf{{x}}^{i}_g &= \mathbf{x}^{i-1} + \text{Attention}(\mathbf{x}^{i-1}+\mathbf{d}_w, \mathbf{p}^{i}+\mathbf{d}_{d}, \mathbf{p}^{i}), \\
    \mathbf{\hat{x}}^{i} &= \mathbf{x}^{i}_g + \text{MLP}(\mathbf{{x}}^{i}_g),
\end{split}
\end{equation}
where $\mathbf{d}_{d} \in \mathbb{R}^{N\times C}$ is the positional encoding for parts in decoders, the definitions of the attention mechanism and MLP are identical to those defined in Eq. \eqref{eq:attn_def} \eqref{eq:rpn_mlp}. Note that $\mathbf{d}_{d}$ is shared across all blocks of each stage.  The axis that the softmax function normalizes on is the last dimension (specifically, the part dimension with $N$ inputs).

\textbf{Patch-based local attention.} The above process, in both the encoder and the decoder, has completed the cross-level interactions for the $i^{th}$ iteration. In addition to the long-range modeling that the cross-level interaction provided, we also apply a local attention for fine-grained feature modeling. 
We divide the spatial feature maps into non-overlapping patches with size $k\times k$, then apply a multi-head self-attention module for all pixels within each patch. We denote the pixels of patch $t$ as $\mathbf{x}^{i}_t \in \mathbb{R}^{k^2 \times C}$, then the process of the local attention can be written as:
\begin{equation}
\begin{split}
    \mathbf{\hat{x}}^{i}_t &= \mathbf{x}^{i}_t + \text{Attention}(\mathbf{x}^{i}_t, \mathbf{x}^{i}_t+\mathbf{r}^i, \mathbf{x}^{i}_t), \\
    \mathbf{\hat{x}}^{i}_l &= \{\mathbf{\hat{x}}^{i}_1, ..., \mathbf{\hat{x}}^{i}_t, ..., \mathbf{\hat{x}}^{i}_{N_p}\},\\
    \mathbf{x}^{i} &= \mathbf{\hat{x}}^{i}_l + \text{MLP}(\mathbf{\hat{x}}^{i}_l),
\end{split}
\end{equation}
where $N_p = \frac{L}{k^2}$ denotes the total number of patches, $\mathbf{r}^i \in \mathbb{R}^{k^{2}\times C}$ is the relative positional embedding. The implementation of the relative positional embedding follows the design in \citep{aa, botnet}. To save the computational cost, $\mathbf{r}^i$ is factorized into two embeddings $\mathbf{r}^i_h \in \mathbb{R}^{(2k-1)\times \frac{C}{2}}, \mathbf{r}^i_w \in \mathbb{R}^{(2k-1)\times \frac{C}{2}}$ for the dimension of height and width respectively.

\subsection{Architecture Specification.}
\label{sec:arch}
In this paper, we design five kinds of different variants called ViP-Mobile (Mo), ViP-Tiny (Ti), ViP-Small (S), ViP-Medium (M), ViP-Base (B) respectively.
These variants have some common features in design.
(1) Architectures of all these models are divided into four stages according to the spatial resolution $L$ of the feature map. Given an input with spatial size $H\times W$, the output spatial sizes of the feature maps for the four stages are $\frac{H}{4}\times\frac{W}{4}$, $\frac{H}{8}\times\frac{W}{8}$, $\frac{H}{16}\times\frac{W}{16}$ and  $\frac{H}{32}\times\frac{W}{32}$.
(2) The expansion rates of the MLP within the encoder and the decoder, which indicates the ratio between the number of channels of the hidden output and the input, are set to be 1 and 3 separately. 
(3) The patch size for the self-attention module of the decoder is set to $\{8, 7, 7, 7\}$ for four different stages.
(4) At the beginning of each stage, there is a patch embedding responsible for down-sampling and channel-wise expansion. We employ a separable convolution with normalization here with kernel size $3\times 3$ to perform the down-sampling operation for the whole representation. Since the number of channels of the part representation for each stage may vary, another linear operation is applied to align the number of channels between parts in different stages.

Apart from these common hyper-parameters, for a specific stage $s$, these variants mainly differ in the following aspects:
(1) The number of channels $C_s$,
(2) The number of parts $N_s$,
(3) The number of blocks $B_s$,
(4) The number of groups (heads) $G_s$ for the attention mechanism.
(5) For small models ViP-Mo, ViP-Ti and ViP-S, we employ a part encoder on top of the whole representation before the final global pooling and fully connected layer, while for ViP-M and ViP-B, we replace such encoder with a linear layer.
The overall architecture of the ViP family is shown in Table \ref{tab:table_net}. The detailed specification of the four variants can be found in the appendix.

\section{Related Work}
\textbf{Convolutional Neural Networks (CNNs).}
Conventional CNNs \citep{vggnet, googlenet, resnet, densenet, efficientnet, regnet} are prevalent in nearly all fields of computer vision since AlexNet \citep{alexnet} demonstrates its power for image recognition on ImageNet \citep{imagenet_cvpr09}. Now CNNs are still dominating nearly all major tasks in computer vision \textit{e.g.} image classification \citep{nfnet, lambdanet, efficientnet, resnet, regnet,sun2019fishnet}, object detection \citep{girshick2014rcnn,girshick2015fastrcnn,fasterrcnn,liu2016ssd,he2017mask,cascade-rcnn,focal_loss} and semantic segmentation \citep{long2015fully,chen2015semantic,yu2016multi,badrinarayanan2017segnet,zhao2017pspnet}.

\textbf{Part-whole hierarchies in visual representations.} 
\citet{tu2005image} first devise a Bayesian framework to parse the image into a part-whole hierarchy for unifying all the major vision tasks.
Capsule Networks (CapsNets) \citep{dynamic_routing} were first proposed to use a dynamic routing algorithm to dynamically allocate neurons to represent a small portion of the visual input. There are some other extensive works based on CapsNets \citep{em_routing, stack_capsule, capsule_dot} showing remarkable performance on some small datasets, however, these works cannot be well scaled onto large datasets. The recent proposal of GLOM \citep{glom} gives an idea to build up a hierarchical representation with attention, but it gives no practical experiments. This paper borrows some ideas from these works to build a rather simple hierarchy with two levels for modeling basic visual representation.
For example, the iterative attention mechanism in our model is similar to the dynamic routing designed in CapsNet \citep{dynamic_routing} or iterative attention mechanism \citep{capsule_dot, slot_attention, detr}.

\textbf{Transformers and self-attention mechanism.}
With the success of Memory Networks \citep{memorynet} and Transformers \citep{transformer} for natural language modeling \citep{wu2019pay,Devlin2018,dai2019transformer,Yang2019xlnet}, lots of works in the field of computer vision attempted to migrate similar self-attention mechanism as an independent block into CNNs for image classification \citep{senet, glore, aa, visualtrans}, object detection \citep{hu2018relationnet, botnet, detr} and video action recognition \citep{jia2016dynamic, nonlocal,multiscale_trans}.

Recent works tried to replace all convolutional layers in neural networks with local attention layers to build up self-attention-based networks \citep{hu2018relationnet, sasa, san, cordonnier2019relationship, axial_attn, halonet}. To resolve the inefficiency problem, Vision Transformer (ViT) \citep{vit, deit} chose to largely reduce the image resolution and only retain the global visual tokens while processing. To aid the global token-based transformer with local inductive biases, there are several papers that incorporate convolution-like design into ViT \citep{t2t, cpvt, d2021convit, pit, cvt}. Apart from the token-based approach, concurrent works \citep{swin, pvt} that retain the spatial pyramids has also been proven to be effective. Different from the above existing works that extract either tokens or spatial feature maps for final prediction, ViP extracts both the token-based representations (the part) and spatial feature maps (the whole).

\textbf{Token-based global attention mechanism.} The interaction between the part and the whole is related to the token-based global attention mechanism. Recent works including \citep{glore, lambdanet, a2net, visualtrans} propose to tokenize the input feature map generated by the convolution block. Our work is different from theirs in three aspects:
(1) The part representations are explicitly and iteratively refined in ViP, while in these works, the tokens are latent bi-product of each block.
(2) We intend to design a hierarchy that can be used for final prediction while these works focus on designing a module then plug it into limited blocks of the network.
(3) In detail, the token extraction and the bi-directional pathway designed in ViP are quite different from their pipelines.

\begin{table}[t]
\begin{minipage}{\textwidth}
\begin{minipage}[t]{0.48\textwidth}
\strut\vspace*{-\baselineskip}\newline  
    \setlength{\tabcolsep}{1.6pt}
    \centering
    \scalebox{0.8}{
    \begin{tabular}{l|c|c|c|c}
        \Xhline{3\arrayrulewidth}
         Model & \makecell[c]{Input\\Size} & \makecell[c]{Params\\(M)} & \makecell{FLOPS \\(G)} & \makecell[c]{Top-1 \\Acc (\%)} \\
        \Xhline{3\arrayrulewidth}
        \multicolumn{5}{c}{CNN Architectures} \\
        \hline
         BoTNet-T3 \cite{botnet} & 224$^2$ & 33.5 & 7.3 & 81.7  \\
         BoTNet-T4 \cite{botnet} & 224$^2$ & 54.7 & 10.9 & 82.8 \\
         BoTNet-T5 \cite{botnet} & 256$^2$ & 75.1 & 19.3 & 83.5 \\
         \hline
         RegNetY-4G \cite{regnet} & 224$^2$ & 20.6 & 4.0 & 80.0 \\
         RegNetY-8G \cite{regnet} & 224$^2$ & 39.2 & 8.0 & 81.7 \\
         RegNetY-16G \cite{regnet} & 224$^2$& 83.6 & 15.9 & 82.9 \\
         \Xhline{3\arrayrulewidth}
         \multicolumn{5}{c}{Transformer Architectures} \\
         \hline
         ViT-B \cite{vit} & 384$^2$ & 86.4 & 55.4 & 77.9 \\
         ViT-L \cite{deit} & 384$^2$ & 307 & 190.7 & 76.5 \\
         \hline
         DeiT-Ti \cite{deit} & 224$^2$ & 5.7 & 1.6 & 72.2 \\
         DeiT-S \cite{deit} & 224$^2$ & 22.1 & 4.6 & 79.8 \\
         DeiT-B \cite{deit} & 224$^2$ & 86.6 & 17.6 & 81.8 \\
         DeiT-B$\uparrow$384 \cite{deit} & 384$^2$ & 86.6 & 55.4 & 83.1 \\
         \hline
         PVT-Tiny \cite{pvt} & 224$^2$ & 13.2 & 1.9 & 75.1 \\
         PVT-Small \cite{pvt} & 224$^2$ & 24.5 & 3.8 & 79.8 \\
         PVT-Medium \cite{pvt} & 224$^2$ & 44.2 & 6.7 & 81.2 \\
         PVT-Large \cite{pvt} & 224$^2$ & 61.4 & 9.8 & 81.7 \\
         \hline
         T2T-ViT-14 \cite{t2t} & 224$^2$ & 21.5 & 5.2 & 81.5 \\
         T2T-ViT-19 \cite{t2t} & 224$^2$ & 39.2 & 8.9 & 81.9 \\
         T2T-ViT-24 \cite{t2t} & 224$^2$ & 64.1 & 14.1 & 82.3 \\
         \hline
         TNT-S \cite{tnt} & 224$^2$ & 23.8 & 5.2 & 81.5 \\
         TNT-B \cite{tnt} & 224$^2$ & 65.6 & 14.1 & 82.9 \\
         \hline
         Swin-T \cite{swin} & 224$^2$ & 29 & 4.5 & 81.3 \\
         Swin-S \cite{swin} & 224$^2$ & 50 & 8.7 & 83.0 \\
         Swin-B \cite{swin} & 224$^2$ & 88 & 15.4 & 83.3 \\
         \hline
         \rowcolor{Gray!16}
         ViP-Mo & 224$^2$ & 5.3 & 0.8 & 75.1 \\
         \rowcolor{Gray!16}
         ViP-Ti & 224$^2$ & 12.8 & 1.7 & 79.0 \\
         \rowcolor{Gray!16}
         ViP-S & 224$^2$ & 32.1 & 4.5 & 81.9 \\
         \rowcolor{Gray!16}
         ViP-M & 224$^2$ & 49.6 & 8.0 & 83.3 \\
         \rowcolor{Gray!16}
         {ViP-B} & 224$^2$ & 87.8 & 15.0 & {83.8} \\
         \rowcolor{Gray!16}
         {ViP-B$\uparrow$384} & 384$^2$ & 87.8 & 39.1 & {84.2} \\
        \Xhline{3\arrayrulewidth}
    \end{tabular}
}
\caption{Results on ImageNet-1K.}
\label{table:imagenet_final}
\end{minipage}
\hfill
\begin{minipage}[t]{0.48\textwidth}
    \begin{minipage}[t]{\textwidth}
    \centering
    \strut\vspace*{-\baselineskip}\newline  
    \includegraphics[scale=0.55]{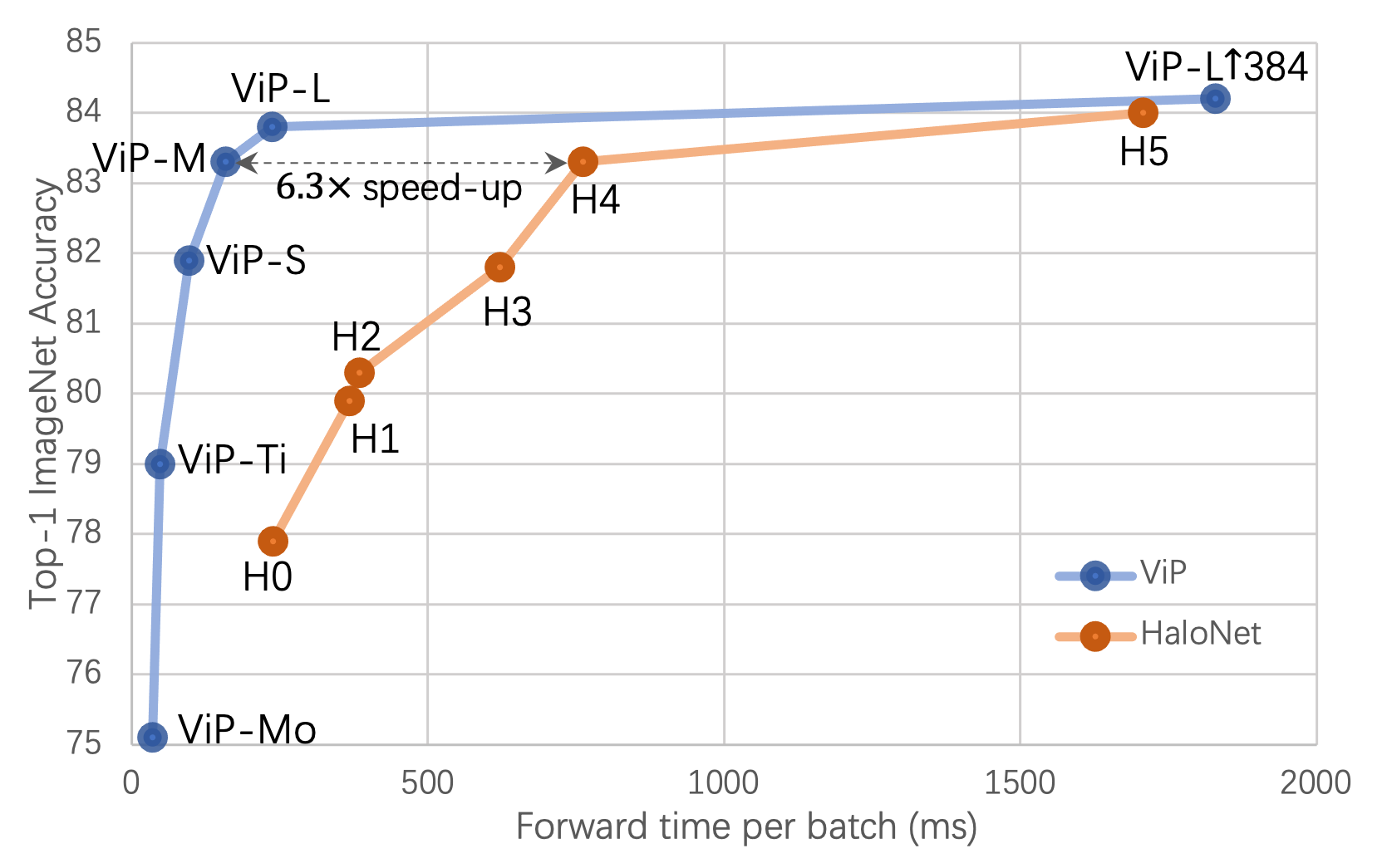}
    \captionof{figure}{\small Speed-Accuracy comparison with HaloNet.}
    \label{fig:halonet}
    \end{minipage}
    
    \begin{minipage}[t]{\textwidth}
    \begin{minipage}{1.0\textwidth}
    \centering
    \setlength{\tabcolsep}{1.6pt}
    \scalebox{0.96}{
    \begin{tabular}{cccc}
    \Xhline{1.5pt}
         & $N$ & \makecell[c]{FLOPS \\ (G)} & \makecell[c]{Top-1 \\ (\%)}\\
         \hline
         \multirow{4}{*}{ViP-Ti} & 8 & 1.6 & 77.6 \\
         & 16 & 1.6 & 78.1 \\
         & 32 & 1.7 & 79.0 \\
         & 64 & 1.8 & 79.1 \\
        \Xhline{1.5pt}
    \end{tabular}
    } %
    \caption{Effect of number of parts for ViP-Ti.} %
    \label{tab:parts}
    \end{minipage} %
    \hfill
    \begin{minipage}{1.0\textwidth}
    \centering
    \setlength{\tabcolsep}{1.6pt}
    \scalebox{0.93}{
    \begin{tabular}{cccc}
    \Xhline{1.5pt}
         & \makecell[c]{Predict\\on parts} & \makecell[c]{Predict\\on wholes} & \makecell[c]{Top-1 \\ (\%)}\\
         \hline
         \multirow{2}{*}{ViP-Ti} & \checkmark & & \textbf{79.0}  \\
         & & \checkmark & 78.3 \\
         \hline
         \multirow{2}{*}{ViP-S} & \checkmark & & \textbf{81.9} \\
         & & \checkmark & 81.5 \\
         \hline
         \multirow{2}{*}{ViP-M} & \checkmark & & 82.7 \\
         & & \checkmark & \textbf{83.3} \\
        \Xhline{1.5pt}
    \end{tabular}
    }
    \caption{Effect of predicting on part/whole level.}
    \label{tab:last_level}
    \end{minipage}
    \end{minipage}
\end{minipage}
\end{minipage}
\vspace{-16px}
\end{table}
\section{Experiments}
\label{sec:experiments}
\subsection{Image Classification on ImageNet-1K}
\textbf{Experimental Settings.} For image classification, we evaluate our models on ImageNet-1K \citep{imagenet_cvpr09}, which consists of 1.28M training images and 50K validation images categorized into 1,000 classes.
The network is trained for 300 epochs using AdamW \citep{adam} and a half-cosine annealing learning rate scheduler. The learning rate is warmed up for 20 epochs to reach the initial $1\times 10^{-3}$. Weight decays for {ViP-Mo,} ViP-Ti are set to be 0.03, while those for ViP-S, ViP-M are 0.05. The drop ratios of Stochastic Depth (\textit{a.k.a} DropPath) \citep{droppath} are linearly scaled from 0 to 0.1, 0.1, 0.2, 0.3 along the layer depth for each layer of ViP-Ti, ViP-S, ViP-M and ViP-B respectively. We do not apply Stochastic Depth to ViP-Mo during training.
The total training batch size is set to be 1024 for all model variants.
We primarily follow the settings of the data augmentation adopted in \citep{deit}, except for the repeat augmentation as we found it unhelpful towards the final prediction. Note that for all results for image classification on ImageNet reported in this paper, we did not use any external dataset for pre-training.

\textbf{ViP \textit{vs.} CNNs.} Table \ref{table:imagenet_final} compares ViP family with some of the state-of-the-art CNN models on ImageNet. 
The RegNet is also better tuned using training tricks in \citep{deit}. ViP is both cost-efficient and parameter-efficient compared to these state-of-the-art models. For example, ViP-M can achieve a competitive $83.3\%$ with only 49.6M parameters and 8.0G FLOPS. The counterpart BOTNet-T5 needs 25.5M more parameters and 11.3G FLOPS to achieve similar performance. {When scaling the input to resolution 384$^2$, ViP-B is able to further improve its top-1 accuracy to $84.2\%$.}

\textbf{ViP \textit{vs.} ViT/DeiT.}
We first compare ViP to the token-based Vision Transformer (ViT), which radically reduces the image resolution at the beginning of the network. When both networks are trained from scratch on the training set of ImageNet-1K, ViP-Ti can outperform ViT-B by 1.1\% with only about $\frac{1}{7}$ of its number of parameters and a fraction of its FLOPS. 
 
 Another token-based vision transformer, DeiT, is also listed in Table \ref{table:imagenet_final} for comparison.
 The basic structure of DeiT is identical to what was proposed in ViT but is trained with more data augmentations and regularization techniques.
 When compare ViP with DeiT, we observe that ViP-Ti can surpass DeiT-Tiny by a significant \textbf{$6.8\%$}.
 As for small models like ViP-S, it can outperform DeiT-S by $2.1\%$, which is even better than a way larger variant DeiT-B of the DeiT family. The remarkable improvement on ImageNet demonstrates the importance of retaining the local features within the network for image recognition.

\textbf{ViP \textit{vs.} HaloNet.}
To the best of our knowledge, HaloNet \citep{halonet} is the current state-of-the-art network on ImageNet-1K. Figure \ref{fig:halonet} shows the speed-accuracy Pareto curve of the ViP family compared to the HaloNet family. Note that the HaloNet is re-implemented by us on PyTorch. As shown in Figure \ref{fig:halonet}, ViP achieves better speed-accuracy trade-off than HaloNet. Concretely, ViP-S is 6.3$\times$ faster than HaloNet-H4 with similar top-1 accuracy on ImageNet-1K.

\textbf{ViP \textit{vs.} other state-of-the-art Transformers.}
As shown in Table \ref{table:imagenet_final}, ViP consistently outperforms previous state-of-the-art Transformer-based models in terms of accuracy and model size.
Especially, ViP-B achieves $83.8\%$ ImageNet top-1 accuracy, which is $0.5\%$ higher than Swin-B \cite{swin} with fewer parameters and FLOPS. A similar trend can also be observed when scaled onto larger models, \textit{e.g.} ViP-M achieves $83.3\%$ top-1 accuracy, outperforming TNT-B \cite{tnt}, T2T-ViT-24 \cite{t2t}, PVT-Large \cite{pvt} by $0.4\%$, $1.0\%$, $1.6\%$ respectively.

\begin{table}[t]
    \setlength{\tabcolsep}{0.8pt}
    \centering
    \scalebox{0.86}{
    \begin{tabular}{l | l c c | c c c | l c c | c c c | c c c c}
    \Xhline{3\arrayrulewidth}
        & \multicolumn{6}{c|}{RetinaNet $1\times$} & \multicolumn{6}{c|}{RetinaNet $3\times$} & & & & \\
        \cline{2-13}
         Backbone & AP$^{b}$ & AP$^{b}_{50}$ & AP$^{b}_{75}$ & AP$^{b}_{S}$ & AP$^{b}_{M}$ & AP$^{b}_{L}$ & AP$^{b}$ & AP$^{b}_{50}$ & AP$^{b}_{75}$ & AP$^{b}_{S}$ & AP$^{b}_{M}$ & AP$^{b}_{L}$ & \multicolumn{2}{c}{\makecell[c]{Params\\(M)}} & \multicolumn{2}{c}{\makecell[c]{\makecell{FLOPS \\(G)}}}\\
         \rowcolor{Gray!16}
         ViP-Mo & \textbf{36.5} & 56.7 & 38.6 & 23.4 & 39.7 & 48.4 & \textbf{39.2} & 59.7 & 41.4 & 25.5 & 42.3 & 51.7 & 5.3 & \ignore{15.2} & 15 & \ignore{166} \\
        \hline
        R18 \citep{focal_loss} & 31.8 & 49.6 & 33.6 & 16.3 & 34.3 & 43.2 & 35.4 & 53.9 & 37.6 & 19.5 & 38.2 & 46.8 & 11.0 & \ignore{21.3} & 37 & \ignore{189} \\
        PVT-T \citep{pvt} & 36.7(+4.9) & 56.9 & 38.9 & 22.6 & 38.8 & 50.0 & 39.4(+4.0) & 59.8 & 42.0 & 25.5 & 42.0 & 52.1 & 12.3 & \ignore{23.0} & 70 & \ignore{221}\\
        \rowcolor{Gray!16}
        ViP-Ti & \textbf{39.7}\gain{7.9} & 60.6 & 42.2 & 23.9 & 42.9 & 53.0 & \textbf{41.6}\gain{6.2} & 62.6 & 44.0 & 27.2 & 45.1 & 54.2 & 11.2 & \ignore{21.4} & 29 & \ignore{181} \\
        \hline
        R50 \citep{focal_loss} & 36.5 & 55.4 & 39.1 & 20.4 & 40.3 & 48.1 & 39.0 & 58.4 & 41.8 & 22.4 & 42.8 & 51.6 & 23.3 & \ignore{37.7} & 84 & \ignore{239} \\
        PVT-S \citep{pvt} & 40.4(+3.9) & 61.3 & 43.0 & 25.0 & 42.9 & 55.7 & 42.2(+3.2) & 62.7 & 45.0 & 26.2 & 45.2 & 57.2 & 23.6 & \ignore{34.2} & 134 & \ignore{286} \\
        \rowcolor{Gray!16}
        ViP-S & \textbf{43.0}\gain{6.5} & 64.0 & 45.9 & 28.9 & 46.7 & 56.3 & \textbf{44.0}\gain{5.0} & 65.1 & 47.2 & 28.8 & 47.3 & 57.2 & 29.0 & \ignore{39.9} & 75 & \ignore{227} \\
        \hline
        R101 \citep{focal_loss} & 38.5 & 57.8 & 41.2 & 21.4 & 42.6 & 51.1 & 40.9  & 60.1 & 44.0 & 23.7 & 45.0 & 53.8 & 42.3 & \ignore{56.7} & 160 & \ignore{315} \\
        X101-32 \citep{focal_loss} & 39.9(+1.4) & 59.6 & 42.7 & 22.3 & 44.2 & 52.5 & 41.4(+0.5) & 61.0 & 44.3 & 23.9 & 45.5 & 53.7 & 41.9 & \ignore{56.4} & 164 & \ignore{319} \\
        PVT-M \citep{pvt} & 41.9(+3.4) & 63.1 & 44.3 & 25.0 & 44.9 & 57.6 & 43.2(+2.3) & 63.8 & 46.1 & 27.3 & 46.3 & 58.9 & 43.7 & \ignore{54.3} & 222 & \ignore{374} \\
        \rowcolor{Gray!16}
        ViP-M & \textbf{44.3}\gain{5.8} & 65.9 & 47.4 & 30.7 & 48.0 & 57.9 & \textbf{45.3}\gain{4.4} & 66.4 & 48.5 & 29.7 & 48.6 & 59.3 & 48.8 & \ignore{59.8} & 135 & \ignore{287} \\
        \hline
        X101-64 \citep{focal_loss} & 41.0 & 60.9 & 44.0 & 23.9 & 45.2 & 54.0 & 41.8 & 61.5 & 44.4 & 25.2 & 45.4 & 54.6 & 81.0 & \ignore{95.5} & 317 & \ignore{473} \\
        PVT-L \citep{pvt} & 42.6 & 63.7 & 45.4 & 25.8 & 46.0 & 58.4 & 43.4 & 63.6 & 46.1 & 26.1 & 46.0 & 59.5 & 60.9 & \ignore{71.5} & 324 & \ignore{476} \\
        
    \Xhline{3\arrayrulewidth}
    \end{tabular}
    }           
    \caption{Various backbones with RetinaNet. Here R and X are abbreviations for ResNet and ResNeXt. Parameters and FLOPS in black are for backbones, while those in \ignore{gray} are for the whole frameworks.}
    \label{tab:retinanet}
\end{table}

\begin{table}[H]
    \setlength{\tabcolsep}{0.8pt}
    \centering
    \scalebox{0.88}{
    \begin{tabular}{l | l c c | l c c | l c c | c c c | c c c c}
    \Xhline{3\arrayrulewidth}
         Backbone & AP$^{b}$ & AP$^{b}_{50}$ & AP$^{b}_{75}$ & AP$^{b}_{S}$ & AP$^{b}_{M}$ & AP$^{b}_{L}$ & AP$^{m}$ & AP$^{m}_{50}$ & AP$^{m}_{75}$ &  AP$^{m}_{S}$ & AP$^{m}_{M}$ & AP$^{m}_{L}$ & \multicolumn{2}{c}{\makecell[c]{Params\\(M)}} & \multicolumn{2}{c}{\makecell[c]{\makecell{FLOPS \\(G)}}}\\
         \rowcolor{Gray!16}
        ViP-Mo & \textbf{42.6} & 61.8 & 46.0 & 27.1 & 44.9 & 56.4 & \textbf{37.7} & 59.2 & 40.1 & 22.5 & 39.8 & 51.0 & 5.9 & \ignore{63.9} & 16 & \ignore{665} \\
        \hline
        R18 & 38.7 & 56.2 & 41.3 & 21.3 & 40.8 & 52.9 & 34.0 & 53.5 & 36.4 & 17.4 & 36.0 & 48.1 &  11.0 & \ignore{69.0}  & 37 & \ignore{686} \\
        \rowcolor{Gray!16}
        ViP-Ti & \textbf{45.4}\gain{6.7} & 64.6 & 48.9 & 29.1 & 48.8 & 60.1 & \textbf{39.9}\gain{5.9} & 61.7 & 42.7 & 24.1 & 43.0 & 54.3 & 11.2 & \ignore{69.2} & \textbf{29} & \ignore{678} \\
        \hline
        R50 & 41.2 & 59.4 & 45.0 & 23.9 & 44.2 & 54.4 & 35.9 & 56.6 & 38.4 & 19.4 & 38.5 & 49.3 & 23.3 & \ignore{82.0} & 84 & \ignore{739} \\
        S50 & 45.4 & 64.1 & 49.2 & 28.3 & 49.1 & 58.8 & 39.5 & 61.4 & 42.5 & 23.1 & 43.0 & 52.8 & 25.1 & \ignore{82.9} & 110 & \ignore{763} \\
        \rowcolor{Gray!16}
        ViP-S & \textbf{48.5}\gain{7.3} & 67.5 & 52.5 & 31.9 & 51.8 & 63.2 & \textbf{42.2}\gain{6.3} & 64.8 & 45.7 & 25.9 & 45.5 & 56.7 & 29.0 & \ignore{87.1} & \textbf{75} & \ignore{725} \\
        \hline
        R101 & 42.9 & 61.0 & 46.6 & 24.4 & 46.5 & 57.0 & 37.3 & 58.2 & 40.1 & 19.7 & 40.6 & 51.5 & 42.3 & \ignore{101.0} & 160 & \ignore{815} \\
        X101-32 & 44.3 & 62.7 & 48.4 & 25.4 & 48.4 & 58.1 & 38.3 & 59.7 & 41.2 & 20.6 & 42.0 & 52.3 & 41.9 & \ignore{100.6} & 164 & \ignore{819} \\
        S101 & 47.7 & 66.4 & 51.9 & 30.1 & 51.8 & 61.4 & 41.4 & 63.7 & 45.1 & 24.7 & 45.2 & 54.9 & 45.7 & \ignore{104} & 209 & \ignore{862} \\
        \rowcolor{Gray!16}
        ViP-M & \textbf{49.9}\gain{7.0} & 69.5 & 54.2 & 33.1 & 53.4 & 65.1 & \textbf{43.5}\gain{6.2} & 66.4 & 47.2 & 26.8 & 46.9 & 59.1 & 48.8 & \ignore{107.0} & \textbf{135} & \ignore{785} \\
        \hline
        X101-64 & 45.3 & 63.9 & 49.6 & 26.7 & 49.4 & 59.9 & 39.2 & 61.1 & 42.2 & 21.6 & 42.8 & 53.7 & 81.0 & \ignore{139.7} & 317 & \ignore{972} \\
    \Xhline{3\arrayrulewidth}
    \end{tabular}
    }
    \caption{Various backbones with Cascade Mask R-CNN. All results are trained under $1\times$ schedule. Here S denotes ResNeSt \cite{zhang2020resnest}.}
    \label{tab:cascade}
\end{table}

\subsection{Object Detection and Instance Segmentation}
\textbf{Experimental settings.} For object detection and instance segmentation, we evaluate ViP on the challenging MS COCO dataset \citep{coco}, which contains 115k images for training (\emph{train-2017}) and 5k images (\emph{val-2017}) for validation. We train models on \emph{train-2017} and report the results on \emph{val-2017}. We measure our results following the official definition of Average Precision (AP) metrics given by MS COCO, which includes AP$_{50}$ and AP$_{75}$ (averaged over IoU thresholds $50$ and $75$) and AP$_{S}$, AP$_{M}$, AP$_{L}$ (AP at scale Small, Medium and Large). Annotations of MS COCO include both bounding boxes and polygon masks for object detection and instance segmentation respectively. 
Experiments are implemented based on the open source mmdetection \citep{mmdetection} platform.
 All models are trained under two different training schedules \emph{$1\times$} (12 epochs) and \emph{$3\times$} (36 epochs) using the AdamW \citep{adam} optimizer with the same weight decay set for image classification. After a 500 iteration's warming-up, the learning rate is initialized at $1\times 10^{-4}$ then decayed by 0.1 after [8, 11] and [27, 33] epochs for \emph{$1\times$} and \emph{$3\times$} respectively. For data augmentation, we only apply random flipping with a probability of 0.5 and scale jittering from 640 to 800. The batch size for each GPU is 2 and we use 8 GPUs to train the network for all experiments. Stochastic Depth is also applied here as what proposed on ImageNet.
 We embed ViP into two popular frameworks for object detection and instance segmentation, RetinaNet \citep{focal_loss} and Cascade Mask-RCNN \citep{cascade-rcnn}.
 When incorporating ViP into these frameworks, ViP serves as the backbone followed by a Feature Pyramid Network (FPN) \cite{lin17fpn} refining the multi-scale whole representations. All weights within the backbone are first pre-trained on ImageNet-1K, while those outside the backbone are initialized with Xavier \citep{xavier}.
 
 \textbf{ViP can outperform ResNe(X)t with $~4\times$ less computational cost in RetinaNet.}
 Table \ref{tab:retinanet} exhibits the experimental results when embedding different backbones into RetinaNet.
 When trained under the $1\times$ schedule, our ViP-Ti can outperform its counterpart ResNet-18 by {$7.9$}, which is a large margin since it is even higher than the performance obtained by ResNet-101 ({$4\times$ larger than ViP-Ti in terms of FLOPS and parameters}). The ViP-S, which is just about the size of ResNet-50, can even outperforms the largest model ResNeXt-101-64$\times$4d listed in Table \ref{tab:retinanet} by a clear {$2.0$}. For larger variant ViP-M, it can further boost the performance to a higher level {$44.3$}.
 The performance of the ViP family can be steadily boosted by the longer $3\times$ training schedule. As shown in Table \ref{tab:retinanet}, all variants of the ViP family can retain their superiority compared with their ResNe(X)t and PVT counterparts.

 \textbf{ViP-Tiny can be comparable with the \emph{largest variant} of ResNe(X)t family in Cascade Mask RCNN.}
 Table \ref{tab:cascade} shows the results when incorporating different backbones into Cascade Mask RCNN \citep{cascade-rcnn}. As shown in Table \ref{tab:cascade}, when trained under $1\times$ schedule, all variants of the ViP family can achieve significantly better performance compared to their counterparts. Notably, as a tiny model with only 11.2M parameters and 29G FLOPS, ViP-Ti can achieve comparable performance obtained by the largest variant in ResNe(X)t family ResNeXt-101-64$\times$4d which contains 81M parameters and 317G FLOPS. ViP also scales well with larger models. ViP-M further lifts the performance to 49.9 for object detection and 43.5 for instance segmentation.
 When compared with state-of-the-art variants of ResNet family like ResNeSt \citep{zhang2020resnest}, ViP can also outperform them by a clear margin. Specifically, ViP-S and ViP-M outperforms ResNeSt-50 and ResNeSt-101 by $3.1$ and $2.2$ respectively on object detection and $2.7$ and $2.1$ on instance segmentation.

\begin{figure}
    \centering
    \def\arraystretch{0.2}
    \begin{tabular}{c@{\hspace{0.2mm}}c@{\hspace{0.2mm}}c}
    Block 3 & Block 4 & Block 5 \\
        {\includegraphics[width=0.33\linewidth]{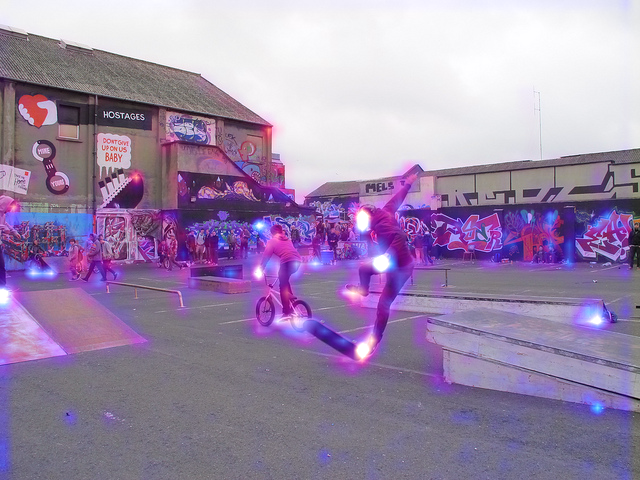}} & {\includegraphics[width=0.33\linewidth]{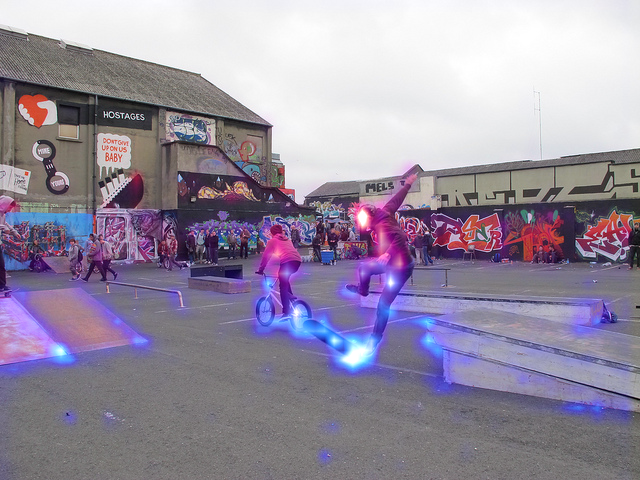}} & {\includegraphics[width=0.33\linewidth]{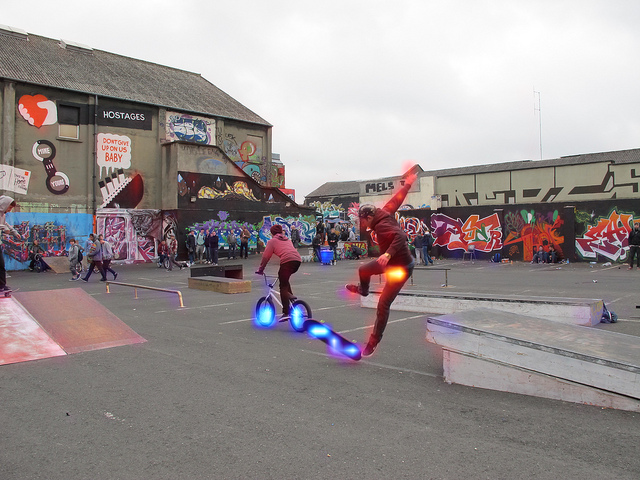}} \\
        {\includegraphics[width=0.33\linewidth]{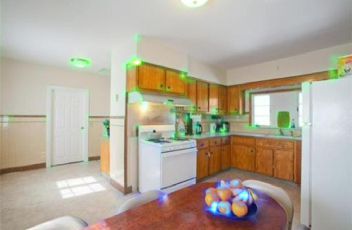}} & {\includegraphics[width=0.33\linewidth]{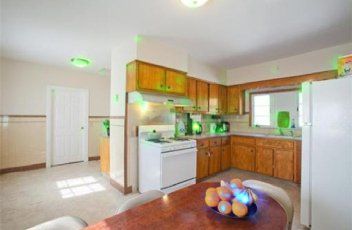}} & {\includegraphics[width=0.33\linewidth]{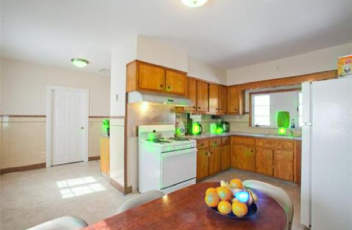}} \\
        {\includegraphics[width=0.33\linewidth]{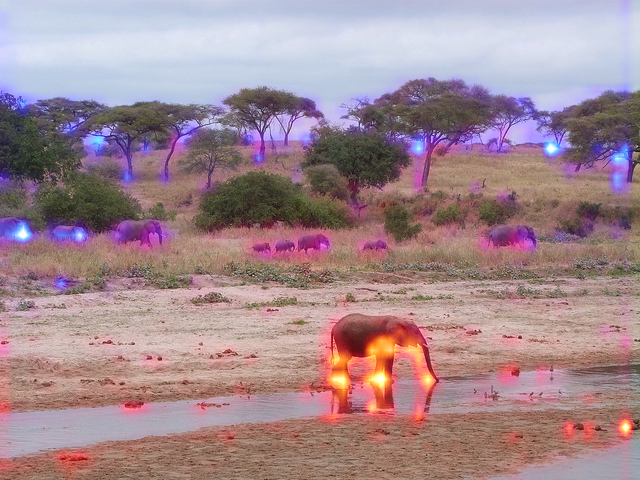}} & {\includegraphics[width=0.33\linewidth]{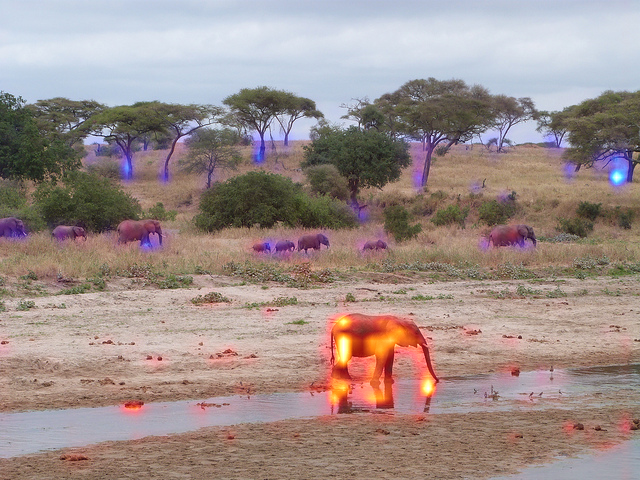}} & {\includegraphics[width=0.33\linewidth]{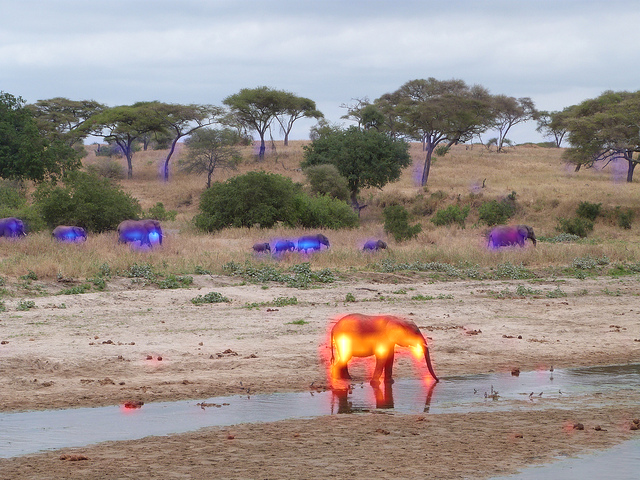}} \\
    \end{tabular}
    \caption{Visualization results about where the part representations attend on. Pixels rendered in different colors are associated to different parts. Best viewed in color.}
    \label{fig:viz}
\end{figure}

\subsection{Ablation studies}
\textbf{Number of parts:} As shown in Table \ref{tab:parts}, the number of parts $N$ is crucial when the model is small. Concretely, for ViP-Ti, $N$=32 can lead to a remarkable $0.8\%$ improvement on ImageNet compared to $N$=16. However, the improvement comes to be saturated when adding more parts into the network.

\textbf{Effects of the part-wise linear.} Different from the original design in Transformer \citep{transformer} that uses a self-attention module for part-wise communication, we replace it with a simple linear operation to save the computational cost. Introducing such a simple linear operation into ViP-S can lead to a $0.4\%$ gain on ImageNet with only a fractional increase in parameters (0.03M) and FLOPS (0.02G).

\textbf{Predicting on parts \textit{vs.} wholes.} As ViP has two levels, we can choose either of them for the final prediction. When using the parts for the final prediction, we employ an additional encoder on top of the whole before the final global pooling and fully connected layer to gather all parts obtained by the encoder. Otherwise, we replace the encoder with a linear projection.
Table \ref{tab:last_level} compares the results of predicting on the part representation. For small models like ViP-Ti and ViP-small, to predict on parts can lead to remarkable improvements ($+0.7\%$ and $+0.4\%$). However, predicting on part level encountered an overfitting problem when incorporated into ViP-M with a $0.6\%$ drop.

\subsection{Visualization}
\label{sec:viz}
The visualization results of a pre-trained ViP-S are shown as an example in Figure~\ref{fig:viz}. 
We average all heads of the affinity matrix $\mathbf{M}$ in Eq. \eqref{eq:softmax} and then normalize it to [0, 255]. For each image, we visualize two parts in total for clearness.
It can be observed that the attention maps tend to cover more area in the shallow blocks, and then gradually focus on salient objects through multiple iterations. The observation suggests that the part encoder can effectively aggregate features from a part of the image, and the part representations can be refined with the information from the whole.
The visualization results show that a meaningful part-whole hierarchy is constructed by the proposed ViP.

\section{Conclusion}
In this work, we construct a framework that includes different levels of representations called Visual Parser (ViP). ViP divides visual representations into the part level and the whole level with a novel encoder-decoder interaction. Extensive experiments demonstrate that the proposed ViP can achieve very competitive results on three major vision tasks. Particularly, ViP outperforms the previous state-of-the-art CNN backbones by a large margin on object detection and instance segmentation.
Visualization results also indicate that the learned part representations are highly informative to the predicting classes. As the first step towards learning multi-level part-whole representations, our ViP is more explainable compared to previous architectures and shows great potential in visual modeling. 

\noindent \textbf{Acknowledgement.} 
We would like to thank Pau de Jorge, Francesco Pinto, Hengshuang Zhao and Xiaojuan Qi for proof-reading and helpful comments. 
This work is supported by the ERC grant ERC-2012-AdG 321162-HELIOS, EPSRC grant Seebibyte EP/M013774/1 and EPSRC/MURI grant EP/N019474/1. We would also like to thank the Royal Academy of Engineering and FiveAI.

\section{Appendix}

\subsection{Network Specification of ViP}
\begin{table}[h]
    \setlength{\tabcolsep}{1.6pt}
        \renewcommand{\arraystretch}{1.5}
        \centering
        \scalebox{0.76}{
        \begin{tabular}{c|c|c|c|c|c|c}
        \Xhline{1.5pt}
            Stage & L 
            & \makecell[c]{ViP-Mobile} 
            & \makecell[c]{ViP-Tiny} 
            & \makecell[c]{ViP-Small} 
            & \makecell[c]{ViP-Medium}
            & \makecell[c]{ViP-Base}\\
            \hline
             & & \multicolumn{5}{c}{$7\times7, 64$, Conv, stride 2;\quad $3\times 3$ Max Pool, stride 2}
             \\
            \hline
            \multirow{3}{*}{Stage$_1$} & \multirow{3}{*}{$\frac{\text{HW}}{4^2}$}
            & Patch Embedding, $48$ 
            & Patch Embedding, $64$ 
            & Patch Embedding, $96$ 
            & Patch Embedding, $96$ 
            & Patch Embedding, $128$ 
            \\ \cline{3-7}
            & & \makecell[c]{
            $\begin{bmatrix} C_1=48\\ N_1=16\\ G_1=1  \end{bmatrix} \times 1$}
            & \makecell[c]{
            $\begin{bmatrix} C_1=64\\ N_1=32\\ G_1=1  \end{bmatrix} \times 1$}
            & \makecell[c]{
            $\begin{bmatrix} C_1=96\\ N_1=64\\ G_1=1  \end{bmatrix} \times 1$}
            & \makecell[c]{
            $\begin{bmatrix} C_1=96\\ N_1=64\\ G_1=1  \end{bmatrix} \times 1$}
            & \makecell[c]{
            $\begin{bmatrix} C_1=128\\ N_1=64\\ G_1=1  \end{bmatrix} \times 1$}\\
            \hline
            \multirow{3}{*}{Stage$_2$} & \multirow{3}{*}{$\frac{\text{HW}}{8^2}$}
            & Patch Embedding, $96$ 
            & Patch Embedding, $128$ 
            & Patch Embedding, $192$ 
            & Patch Embedding, $192$ 
            & Patch Embedding, $256$ 
            \\ \cline{3-7}
             & & \makecell[c]{$\begin{bmatrix} C_2=96\\ N_2=16\\ G_2=2 \end{bmatrix} \times 1$}
             & \makecell[c]{$\begin{bmatrix} C_2=128\\ N_2=16\\ G_2=2 \end{bmatrix} \times 1$}
             & \makecell[c]{$\begin{bmatrix} C_2=192\\ N_2=16\\ G_2=2 \end{bmatrix} \times 1$}
             & \makecell[c]{$\begin{bmatrix} C_2=192\\ N_2=16\\ G_2=2 \end{bmatrix} \times 1$}
             & \makecell[c]{$\begin{bmatrix} C_2=256\\ N_2=16\\ G_2=2 \end{bmatrix} \times 1$}
             \\
            \hline
            \multirow{3}{*}{Stage$_3$} & \multirow{3}{*}{$\frac{\text{HW}}{16^2}$}
            & Patch Embedding, $192$
            & Patch Embedding, $256$
            & Patch Embedding, $384$
            & Patch Embedding, $384$
            & Patch Embedding, $512$
            \\ \cline{3-7} 
            & & \makecell[c]{$\begin{bmatrix} C_3=192\\ N_3=16\\ G_3=4  \end{bmatrix} \times 1$}
            & \makecell[c]{$\begin{bmatrix} C_3=256\\ N_3=32\\ G_3=4  \end{bmatrix} \times 2$}
            & \makecell[c]{$\begin{bmatrix} C_3=384\\ N_3=64\\ G_3=12  \end{bmatrix} \times 3$}
            & \makecell[c]{$\begin{bmatrix} C_3=384\\ N_3=64\\ G_3=12  \end{bmatrix} \times 8$}
            & \makecell[c]{$\begin{bmatrix} C_3=512\\ N_3=128\\ G_3=16  \end{bmatrix} \times 8$}
            \\
            \hline
            \multirow{3}{*}{Stage$_4$} & \multirow{3}{*}{$\frac{\text{HW}}{32^2}$}
            & Patch Embedding, $384$
            & Patch Embedding, $512$
            & Patch Embedding, $768$
            & Patch Embedding, $768$
            & Patch Embedding, $1024$
            \\ \cline{3-7}
            & & \makecell[c]{
            $\begin{bmatrix} C_4=384\\ N_4=32\\ G_4=8  \end{bmatrix} \times 1$}
            & \makecell[c]{
            $\begin{bmatrix} C_4=512\\ N_4=32\\ G_4=8  \end{bmatrix} \times 1$}
            & \makecell[c]{
            $\begin{bmatrix} C_4=768\\ N_4=64\\ G_4=24  \end{bmatrix} \times 1$}
            & \makecell[c]{
            $\begin{bmatrix} C_4=768\\ N_4=64\\ G_4=24  \end{bmatrix} \times 1$}
            & \makecell[c]{
            $\begin{bmatrix} C_4=1024\\ N_4=128\\ G_4=32  \end{bmatrix} \times 1$}
            \\
            \hline
             & & \makecell[c]{$\begin{pmatrix} C_4=384\\ N_4=32\\ G_4=8 \end{pmatrix}\times 1$}
             & \makecell[c]{$\begin{pmatrix} C_4=512\\ N_4=32\\ G_4=8 \end{pmatrix}\times 1$ }
             & \makecell[c]{$\begin{pmatrix} C_4=768\\ N_4=64\\ G_4=24 \end{pmatrix}\times 1$ }
             & \makecell[c]{$768$, Linear \\ $768$, BatchNorm}
             & \makecell[c]{$1024$, Linear \\ $1024$, BatchNorm}
             \\ \cline{3-7}
             &  & \multicolumn{5}{c}{Global Average Pool;\quad $1000$, Linear}
             \\
            \Xhline{1.5pt}
        \end{tabular}
        }
        \captionof{table}{Detailed specification for ViP family. Contents in $[\cdot]$ represents the basic building block of ViP, and those in $(\cdot)$ are just part encoders. Note that we only apply a part decoder at the end of network to make the final prediction on part level in small models like ViP-Mo/Ti/S.}
    \label{tab:detailed_spec}
\end{table}
The general specification of ViP is illustrated in Section~\ref{sec:arch}. Here in Table \ref{tab:detailed_spec} we show the detailed structure of different variants of the ViP family. Note that for small models including ViP-Mobile, ViP-Tiny, and ViP-Small, we apply another encoder at the end of the network but replacing MLP with the activation function GELU to predicting on the part level. While for larger models ViP-Medium and ViP-Base, we replace the encoder with a linear projection (with Batch Normalization \citep{ioffe2015batchnorm}) so that it can predict on the whole level.

\subsection{More Visualization Results}
\begin{figure}[t]
    \centering
    \def\arraystretch{0.2}
    \begin{tabular}{c@{\hspace{0.2mm}}c@{\hspace{0.2mm}}c}
    Block 3 & Block 4 & Block 5 \\
        {\includegraphics[width=0.33\linewidth]{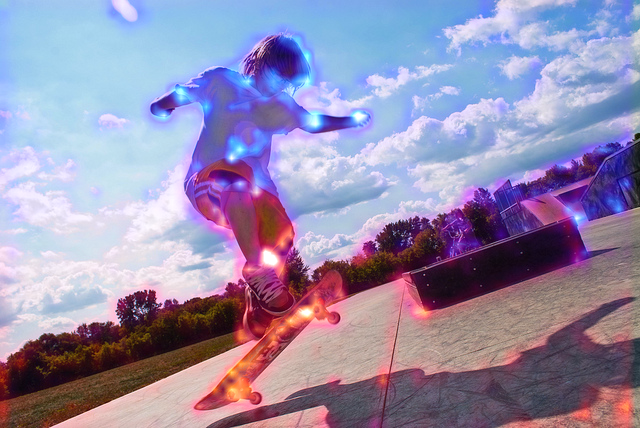}} & {\includegraphics[width=0.33\linewidth]{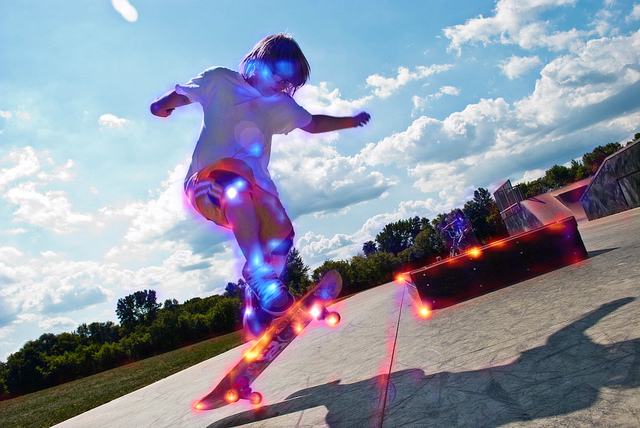}} & {\includegraphics[width=0.33\linewidth]{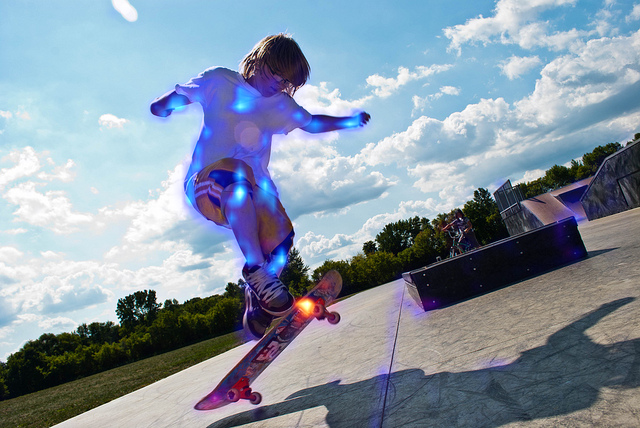}} \\
        {\includegraphics[width=0.33\linewidth]{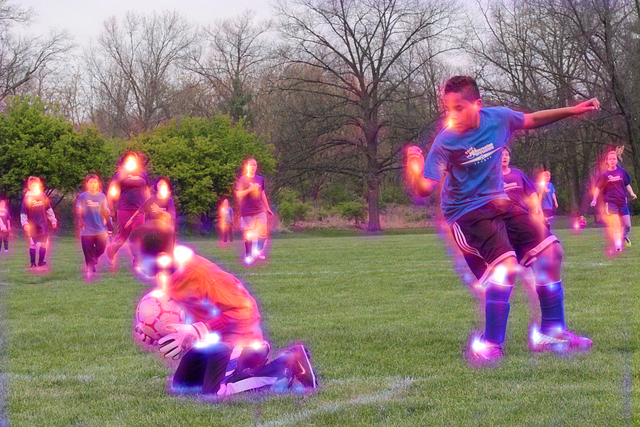}} & {\includegraphics[width=0.33\linewidth]{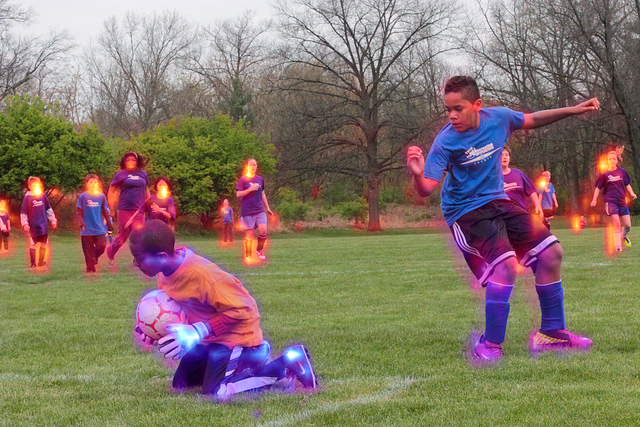}} & {\includegraphics[width=0.33\linewidth]{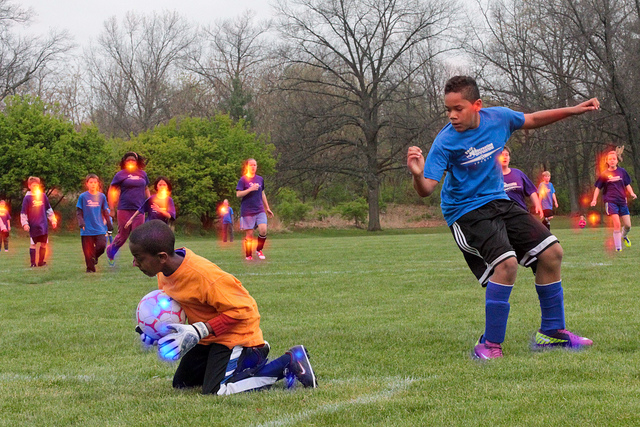}} \\
        {\includegraphics[width=0.33\linewidth]{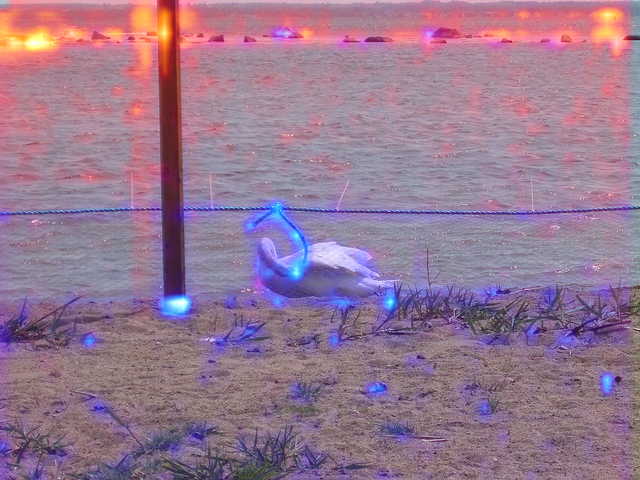}} & {\includegraphics[width=0.33\linewidth]{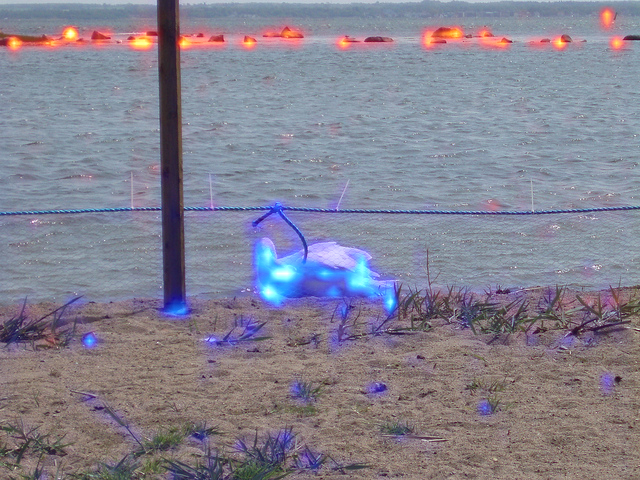}} & {\includegraphics[width=0.33\linewidth]{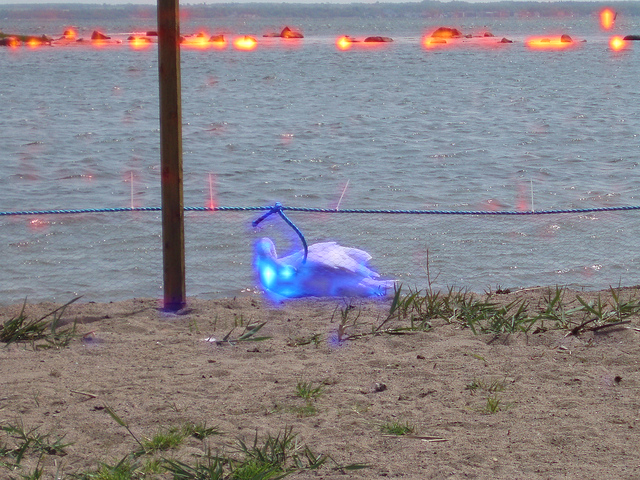}} \\
        {\includegraphics[width=0.33\linewidth]{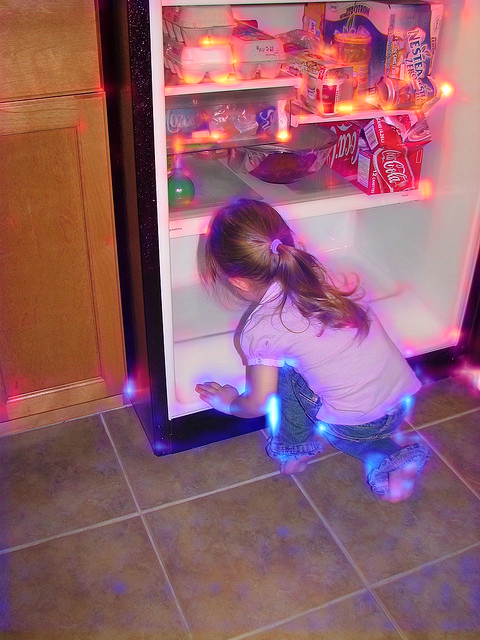}} & {\includegraphics[width=0.33\linewidth]{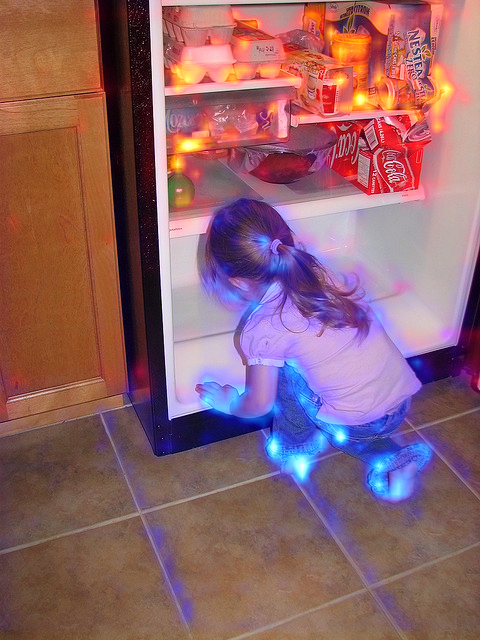}} & {\includegraphics[width=0.33\linewidth]{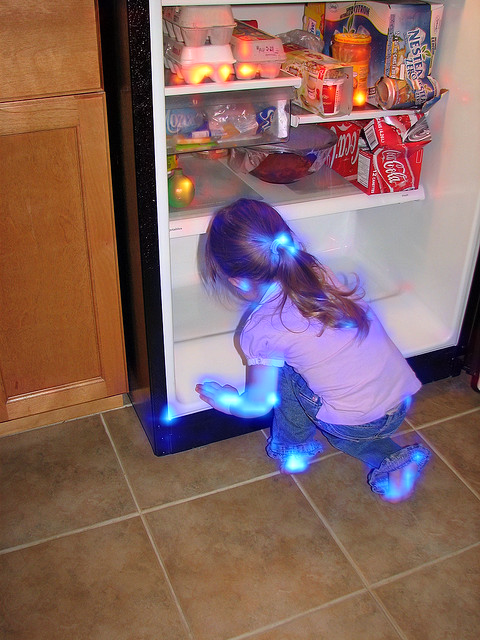}} \\
    \end{tabular}
    \caption{More visualized results. All results are visualized using the same method proposed in Section 4.4.}
    \label{fig:viz_1}
\end{figure}

Additional visualization results are shown in Figure \ref{fig:viz_1}. We follow the visualization method mentioned in Section~\ref{sec:viz}, and it is obvious that the proposed part encoder is robust and works well even for the complex scene.

{
\small
\bibliographystyle{plainnat}
\bibliography{references}
}

\end{document}